# Hybrid Convolution Neural Network Integrated with Pseudo-Newton Boosting for Lumbar Spine Degeneration Detection

Pandiyaraju V[1], Abishek Karthik[2], Jaspin K[3], Kannan A[4], Jaime Lloret[5]

[1, 2]School of Computer Science and Engineering, Vellore Institute of Technology, Chennai, Tamil Nadu, India.

[3]Departmentof Computer Science and Engineering, St. Joseph's Institute of Technology Chennai, Tamil Nadu, India

[4]Department of Information Science and Technology, College of Engineering Guindy, Anna University, Chennai, Tamil Nadu , India

[5]Integrated Management Coastal Zones Research Institute, Universitat Politecnica de Valencia, Spain.

pandiyaraju.v@vit.ac.in,  abishek.sudhirkarthik@gmail.com, jaspink@stjosephstechnology.ac.in , akannan123@gmail.com, jlloret@dcom.upv.es

Corresponding Author: Jaime Lloret, Email: jlloret@dcom.upv.es

*Abstract:*

Lumbar spine degeneration is one of the most prevalent musculoskeletal conditions in older individuals and ranks as a major cause for lower back symptoms, radiculopathy and functional limitation. Effective and early diagnosis is clinically challenging since degenerative phenomena like foraminal narrowing, subarticular stenosis, and spinal canal stenosis present subtle and intricate patterns in MRI/DICOM images. Manual interpretation by a radiologist is labour-intensive and suffers from inter-observer variation, and there is an unmet need for reliable computerized classification tools. In this paper, we present a modified hybrid model for lumbar spine degeneration classification based on DICOM images. The model combines both the VGG19 and EfficientNet as well as two custom-designed layers: Pseudo-Newton Boosting Layer, which incrementally refines feature weights to construe the subtle morphological changes and Sparsity-Induced Feature Reduction (SIFR) layer that discards redundancy in a manner such that it paves way for generating compact yet discriminative features. This full combination can address the drawbacks of conventional transfer learning approaches in high dimensional medical imaging data. Experimental results show that the presented model attains an accuracy of 88.10%, precision of 0.90, recall of 0.86, F1 score of 0.88 and a loss of 0.18 which outperforms other reference models including standalone EfficientNet and VGG19. These findings demonstrate the efficiency of the hybrid strategy for automatic identification of lumbar spine degeneration and represent a direction towards clinically-friendly AI-based diagnostic systems.

# 1. INTRODUCTION

The development of degenerative spinal diseases is an escalating public health problem affecting millions worldwide. More so, this involves lumbar spine degeneration, characterized by both diagnostic and therapeutic complications. This is because it comes along with subtle, complex, and difficult-to-analyze structural changes in spinal discs and vertebrae, which are a challenge to evaluate in a clinical setting on imaging studies. Traditionally, the interpretation was performed by radiologists and orthopedic specialists based on DICOM images [1] with care and attention to spinal anatomy. Nevertheless, variability in the interpretations between physicians and high demand for accuracy necessitate automated systems that would help clinicians generate consistent, high-accuracy image analysis and classification. Recent innovations in smart healthcare solutions have proved the capability of machine learning in disease diagnosis and monitoring, as proved by Rghioui et al. [2], who created a smart healthcare architecture using machine learning algorithms to improve patient monitoring.

Degeneration of the lumbar spine occurs in forms such as foraminal narrowing, subarticular stenosis, and spinal canal stenosis. Foraminal narrowing is defined as a decrease in space within the Foramina, often leading to compression of exiting nerve roots and subsequent pain or weakness radiating to the leg. Subarticular stenosis can manifest as narrowing beneath the facet joint and frequently results in symptoms associated with nerve compression symptoms. Finally, spinal canal stenosis represents narrowing of the central canal that contains the spinal cord, producing neurogenic claudication and potentially severe impairments in function. Epidemiologic studies indicate that lumbar degeneration is remarkably prevalent among older individuals, with potentially up to 9–14% of adults older than 60 years of age exhibiting symptomatic spinal stenosis. The clinical ramifications of these conditions are significant regarding the etiology of lower back pain, radiculopathy, and functional impairment, each with been associated with significant quality of life and healthcare costs. In light of the subtle morphological changes in MRI and DICOM images, it is critical to detect early and accurately to guide intervention and improve patient care and outcomes. These clinical ramifications further contribute to the need for automated deep learning–based approaches that assist radiologists in reliable and consistent classification of lumbar spine degeneration.

Deterioration of the lumbar spine increases the risk of chronic pain development and is one of the most common indications for older adults to undergo spinal surgery, all of which increases healthcare costs. The socioeconomic burden is significant, as spinal degenerative conditions rank among the top disabling conditions in terms of disability adjusted life years (DALYs) worldwide. Early and accurate identification is important because early interventions can slow disease progression, facilitate mobility, and reduce surgical risks. Manual diagnoses are highly subjective because the agreement between radiologists grading severity at the time of diagnosis is notoriously poor and the increasing number of individuals that require spinal imaging has placed stress on healthcare systems even after establishing clinically significant changes. This creates a compelling argument for the development of automated, reliable, and reproducible clinical decision support systems. Artificial Intelligence (AI), especially deep learning methods, suggest the capability of expedient and reliable assessment through the scaling of systems which can process large amounts of medical imaging diagnostic speculations, reduce variability in diagnostics, and increase productivity without sacrificing accuracy.

Advances in machine learning and its specific area of deep learning would allow for the building of automated diagnostic systems, which could potentially be comparable to or even outperform human subjects in specific tasks. Equipped with convolutional neural networks and transfer learning models, these systems have thus far been able to successfully analyse complex medical images-from dermatological through radiological and pathological. For example, Tomás et al. [3] used transfer learning for real-time detection of facemask wearing, highlighting its usefulness in image-based classification tasks. Analogously, García et al. [4] proposed an m-health service for stroke detection based on cloud-based AI-based analysis, supporting the possibility of intelligent medical systems. The ease of leverage with vast amounts of learned information and application to very specified contexts with relatively limited data have popularized transfer learning, especially the use of pre-trained networks, such as EfficientNet and VGG19, fine-tuned on a specific medical dataset. While transfer learning has achieved great success, conventional forms of transfer learning do not seem to be perfectly able to deal with the complex and very subtle degenerative patterns of the lumbar spine in an imaging context, because minute anatomical changes may reflect important clinical implications. Therefore we, introduce a new hybrid model combining a pre-trained EfficientNet and VGG19 architectures supplemented with two novel modules, namely a Pseudo-Newton Boosting layer and a Sparsity-Induced Feature Reduction layer. The Pseudo-Newton Boosting layer is intended to iteratively refine the weights of learned features in order to capture very subtle morphological changes associated with lumbar spine degeneration. This enables adaptively enhancing the weights associated with feature learning and accordingly shifts the emphasis across training epochs toward more pertinent features for improving overall precision in degenerative classification. This will be complemented by the Sparsity-Induced Feature Reduction layer, reducing redundancy by encouraging sparse representations in learned features, ensuring that the model stays computationally efficient and accurate when dealing with high dimensional data. Coupling these architectural enhancements together, the model is now set to transcend these conventional limitations of traditional transfer learning frameworks in a robust solution that is more sensitive to the needs of high-precision spinal imaging diagnostics.

The goals of this work are threefold: It first develops a deep learning model that uses transfer learning but further completes it with specially designed layers specifically for the classification of lumbar spine degeneration. The second challenge is to evaluate the contribution of the Pseudo-Newton Boosting and Sparsity-Induced Feature Reduction layers to both the accuracy of the model and its computational efficiency. Third, it describes the overall performance of the model on the DICOM images in comparison with baseline models, attempting to establish the model's effectiveness as an automated diagnostic tool. In pursuit of the above objectives, this study endeavours to make a meaningful contribution to the field of automated medical diagnostics, potentially offering a valuable tool for the early detection and classification of spinal degeneration. This further enables an illustration of what generalizability of custom layers means in the context of transfer learning frameworks and is highly useful to the design of advanced medical image classification systems with both accuracy and interpretability.

The rest of the paper is organized as follows. Section 2 discusses the current methods and limitations concerning lumbar spine classification. In Section 3, we introduce the proposed hybrid model of data preprocessing and use EfficientNet and VGG19 for feature extraction, followed by applying Pseudo-Newton Boosting to integrate and create the Sparse Induced Feature Reduction layer to achieve better performance. Experimental results in Section 4 compare the new model performance in terms of accuracy, precision, recall, and focal loss with the baseline architectures. This is followed by a wrap-up of this study and future directions on improving model generalization for clinical application in Section 5.

## 2. LITERATURE SURVEY

In the last few years, there has been a considerable advancement in the use of machine learning and deep learning techniques to the analysis of medical images. A diverse scope of methods has been tested to improve diagnostic and classification accuracy, diagnostic efficiency, and diagnostic interpretability (e.g., boosting methods, transfer learning, convolutional neural networks (CNN) approaches, and hybrid/ensemble approaches). However, there are challenges with generalizability, computational cost, and scalability with each method, despite some if not all of these methods demonstrating effective predictive validity. To simply outline a summary of the current state of the art, Table 1 briefly describes the method, findings, and limitations of key contributions from prior studies in regard to lumbar spine degeneration and medical image classification problems.

**Table 1: Comparative Summary of Existing Works in Medical Image Analysis**

| Paper (Citation) | Method(s) | Achievements | Limitations |
|---|---|---|---|
| [5] Yang et al. (2008) | Boosting-based distance metric learning | Improved image retrieval, effective feature learning for varied data | Dependence on static features in pretrained models |
| [6] Ali et al. (2014) | Boosting Stochastic Newton (BSN) | Faster convergence in large image classification tasks | Data heterogeneity challenges in transfer learning |
| [7] Malik et al. (2020) | Transfer Learning | High diagnostic performance | Limited generalization due to dataset regional differences |
| [8] Al-kubaisi et al. (2021) | Transfer Learning for lumbar spine classification | ROI optimization, improved accuracy | Anatomical variation impacts generalization |

| [9] Bayangkari et al. (2023) | EfficientNet + Transfer Learning | Good balance of accuracy and efficiency in classification tasks | Requires large labeled datasets, limiting clinical utility |
|---|---|---|---|
| [10] Miglani et al. (2020) | EfficientNet | Skin lesion classification with high accuracy | Data dependency, computationally heavy |
| [11] Kim et al. (2024); [12] Mantha et al. (2021) | CNN architectures (ResNet50, EfficientNet-B3) | Strong classification performance | High computational resource requirements |
| [13] Han et al. (2021) | VGG19 + Fourier transform zero-watermarking | Robust watermarking to prevent tampering | Computationally expensive |
| [14] Rajinikanth et al. (2020) | VGG19 + handcrafted features ensemble | Improved tumor detection | Overfitting on limited datasets |
| [15] Bharadwaj et al. (2023) | V-Net + Big Transfer (BiT) | High interpretability in spine classification | Accuracy sacrificed for interpretability |
| [17] Truong et al. (2024); [19] Zhang et al. (2023) | Two-layer ensemble; SMOTE-RFE-XGBoost | Balances data imbalance and segmentation accuracy | Extensive hyperparameter tuning makes scaling difficult |

The analysis of the preceding studies reveals various important trends. To start with, boosting-based methods are flexible and efficient, allow for faster convergence, and provide strong feature learning for a wide range of data. Secondly, transfer learning methods have produced impressive results, particularly for EfficientNet and VGG19 models, although the models do not necessarily generalize well, and are limited by the variability of the data used in practice, and across heterogeneous populations. Third, EfficientNet strikes a balance of efficiency and accuracy compared to other models, although like other models, they often require large amounts of annotated data and computational resources, creating limitations for supply. Similarly, CNN models such as ResNet50 or EfficientNet-B3 ultimately performed at a similar predictive capacity but required significantly more memory and computational costs. Fourth, hybrid and ensemble methods really do a good job addressing interpretability and data imbalance but oftentimes lead to models overfitting to a smaller data set and repeatedly propose searching for hyperparameters, again leading to a lack of scalability.

Putting this all together, even though traditional methods have their own advantages, none of them performed well in terms of accuracy, generalizability, efficiency or interpretability simultaneously - demonstrating the possibilities of frameworks to balance competing demands.

To address these gaps, this research proposes a hybrid deep-learning framework based on EfficientNet and VGG19 with the additional contribution of two customized components: a pseudo-Newton boosting layer and a sparsity-induced feature reduction layer. In contrast to conventional transfer learning schemes that shoulder somewhat rigid pre-trained features, our model dynamically modulates the impact of these features to resonate with their changing morphology associated with lumbar degeneration. The Pseudo-Newton Boosting layer highlights feature weights iteratively refining to improve classification accuracy, while the Sparsity-Induced Feature Reduction layer provides sparse optimization by minimizing redundancy, ensuring greater efficiency and generalizability at the same time. In this way, by improved feature selection and a better control of the balance between accuracy and computational efficiency, the method circumvents the disadvantages of transfer learning and ensemble models; in effect, this model delivers a scalable and adaptive one.

## 3. METHODS

In this paper, we propose a novel approach to overcome the deficiencies in conventional pre-trained and single-model architectures towards medical image classification focusing on complex lumbar spine conditions. While most work done so far relies heavily on pre-trained models like VGG and ResNet, our proposed model comes with a hybrid architecture of EfficientNet and VGG19 with augmentation using a Pseudo-Newton Boosting layer and Hybrid Feature Transformer. This framework is designed with enhanced adaptability and accuracy concerning issues of data diversity and dependency that usually dramatically limit generalization ability within traditional models. Our approach eliminates the overfitting that may be caused by introducing the Hybrid Feature Transformer to extract relevant features with efficiency in a diversified range of patient demographics, thereby enhancing performance in multi-modal imaging scenarios.

In addition, we have applied Pseudo-Newton Boosting mechanism that enables fine-tuning and boosting at each step of the metric learning layer to achieve more stable, flexible, and robust representations with a minimal cost. The innovation makes the model more interpretable but is for real-time diagnostics since it helps speed up the convergence process. It is also computationally efficient and scalable, with the solution that is very adaptable with high interpretability, particularly suitable for real-world clinical applications where conventional models may not be applicable or become cumbersome in resource-constrained environments.

To evaluate the performance, there will be an adequate territory to compare various aspects. The model has been developed using a DICOM dataset [1] of the lumbar spine, where the preprocessing procedures include resizing, normalizing, augmenting, and cropping the region of interest (ROI) to highlight the spinal structures. The performance evaluation shall be measured by standard classification metrics: accuracy, precision, recall, F1 score, and loss function-categorical cross-entropy for each epoch. In addition, we shall make use of cross-validation techniques such as k-fold cross-validation and determine AUC-ROC capacity to differentiate the model conditions. To save on computational expense, duration of training and inference, memory consumption, and convergence rate will be evaluated through an overall examination of the loss curves through the epochs. The conclusion of that study will provide insights into the model's scale, accuracy, and applicability in a real-life hospital scenario with low resources.

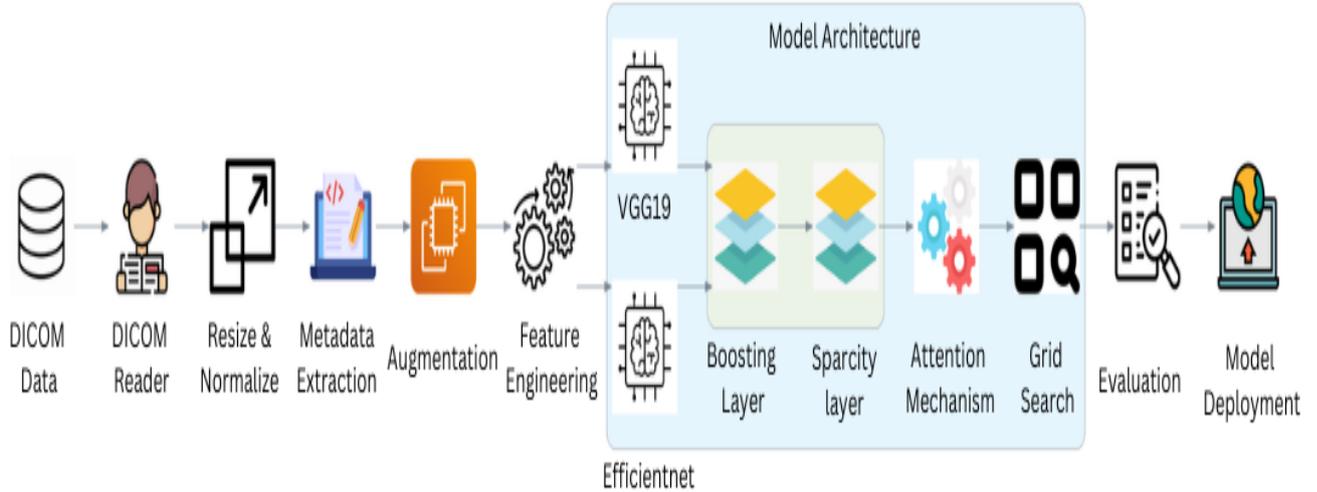

**Figure 1: Overall Proposed System Architecture**

Figure 1 represents the overall architecture which starts with raw DICOM data from lumbar spine imaging, which it reads and then preprocesses for standardization into this model. The first preprocessing steps consist of resizing and normalization, where every pixel intensity $X$. By the formula, $X$ is normalized:

$$X_{\text{norm}} = \frac{X - \mu}{\sigma} \tag{1}$$

Here, $\mu$ is the mean pixel intensity and $\sigma$ stands for the standard deviation; this ensures consistency over inputs and computational efficiency. Metadata extraction then follows. Here, patient demographics as well as scan properties are captured and transferred into the model so that the model's feature set is augmented beyond visual data only. Data augmentation is then used to create different images by including rotation, flips, and brightness adjustments that mimic variation in real life and increase robustness. Feature engineering is, besides, used to focus on those features most relevant for diagnosis-features that are most likely to capture the discernible patterns related to spine degeneration. The core architecture employs the use of a two-pathway structure with two pre-trained models: VGG19 and Efficient Net. The two models possess different strengths in feature extraction. Specifically, VGG19 captures the fine details, whereas Efficient Net results in balanced accuracy and computational efficiency. The combined features from the output of both models are passed through a new pseudo-Newton boosting layer. Here, the layer refines the learning of features without hindering the primary predictive pathways of the original pathways. Weight updating procedure follows the Newton-Raphson

method:

$$\Delta w = -H^{-1}\nabla L \quad (2)$$

where $H$ is the Hessian matrix of the loss function $L$ that capture the curvature of the cost surface, and $\nabla L$ is the gradient. The layer further scales with the adaptive step size $\eta$:

$$w_{t+1} = w_t - \eta H^{-1}\nabla L(w_t) \quad (3)$$

This pseudo-Newton approach improves the learning of models by better learning the weight adjustment and the way it should react to the loss landscape thereby improving feature weighting. A sparsity-induced layer follows this to optimize efficiency. Using $L1$ regularization, it penalizes the absolute sum of weights in the following:

$$L_{\text{sparsity}} = \lambda \sum |w| \quad (4)$$

This is an $L1$ regularization term, where $\lambda$ is a regularization parameter. This layer forces the model to focus on the most relevant features by reducing the unnecessary dimensions of it, thus improving interpretability and computational efficiency. The attention mechanism also focuses the model on clinically significant regions in each image. The computed weights of attention are:

$$\alpha_i = \frac{\exp(e_i)}{\sum_j \exp(e_j)} \quad (5)$$

where $e_i$ is the importance score for region $i$. This way the network can focus on regions that are the most diagnostic of pathology, thereby increasing interpretability and making the focus of the models to match clinical ones. Finally, the hyperparameters are tuned also through grid search and systematically test combinations to deliver an optimal configuration. Precision, recall, and F1-score further validate model efficacy. After optimizing, the model can finally be deployed; this is a streamlined yet robust architecture balancing a certain level of complexity with operational efficiency. The pseudo-Newton boosting, selective sparsity, and attention mechanism within it help resolve the issues faced due to the inadequacies of the previous models, providing interpretability and specificity of features as well as deployability in diagnostic settings.

## 3.1 Dataset Exploration

The dataset incorporated in this work is sourced from the RSNA Lumbar Spine Degeneration Classification dataset [1], which includes roughly 2,000 MRI studies that have been labeled by experts. These studies are labeled by board-certified spine radiologists to determine if degenerative spinal pathologies are present, how severe they are, and where they occur.

Degeneration of the lumbar spine is typically found to be present in three clinically relevant conditions: foraminal narrowing, subarticular stenosis, and central canal stenosis. Foraminal narrowing is defined as a narrowing of the space found within the intervertebral foramina, which can often cause compression of exiting nerve roots leading to radiating pain or weakness in the leg. Subarticular stenosis is a phenomenon that occurs underneath facet joints and can present with nerve impingement symptoms. Central canal stenosis results in narrowing of the spinal canal itself, causing neurogenic claudication and functional limitations. Epidemiological evidence indicates that lumbar degeneration is highly prevalent in the elderly; diagnosed symptomatic spinal stenosis occurs in nearly 9% of adults aged 60+ and is one of the leading causes for loss of mobility and lower back pain for the geriatric population.

The MRI group of images comprises images acquired from both axial and sagittal planes, which provide complementary views. Axial images are important for identifying decompression of lateral recess and neural roots, while sagittal images demonstrate actively involved areas of canal narrowing and degenerative disc changes at varying levels. The MRI group included both T1- and T2-weighted sequences with their own unique accentuated characteristics. In T1 images, fat-rich tissues such as bone marrow are easily identified, while T2 images accentuated fluid-filled structures such as cerebrospinal fluid (CSF) and oedematous changes. Altogether, having both T1 and T2 images makes degeneration severity easier to visualize and helps connect clinical symptoms with imaging findings. Examples of these MRI denominations and their severity ratings are provided in figures 2-5.

**Figure 2: Subarticular Zone (SZ) Stenosis Grading**

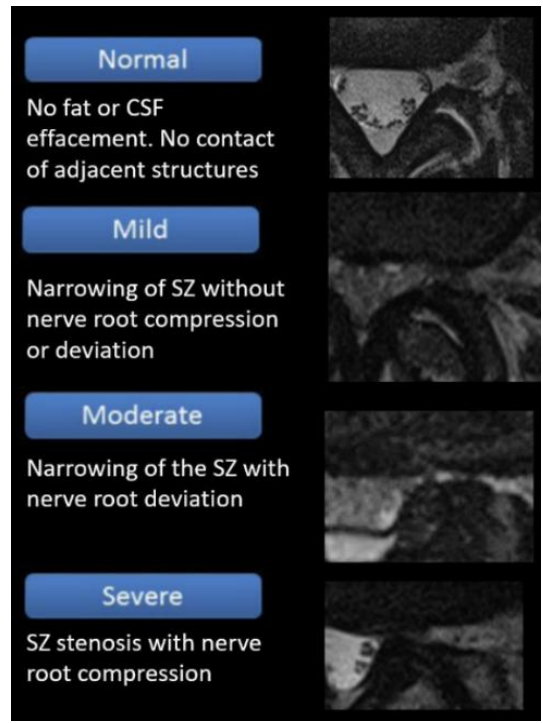

Figure 2 is an example of the severity scoring of lumbar subarticular stenosis on axial MRI slices. The subarticular zone is located adjacent to the vertebral body and pedicles and is frequently colocalized with nerve root compression. Normal images have preserved fat and CSF spaces; mild conditions are defined by the presence of narrowing without deviance; moderate stenosis is characterized by clear nerve root deviance, and severe stenosis indicates significant compression of the nerve root. This visual representation provides corroboration for the severity labels in the current dataset and conforms with the clinical concept of progressive pathology.

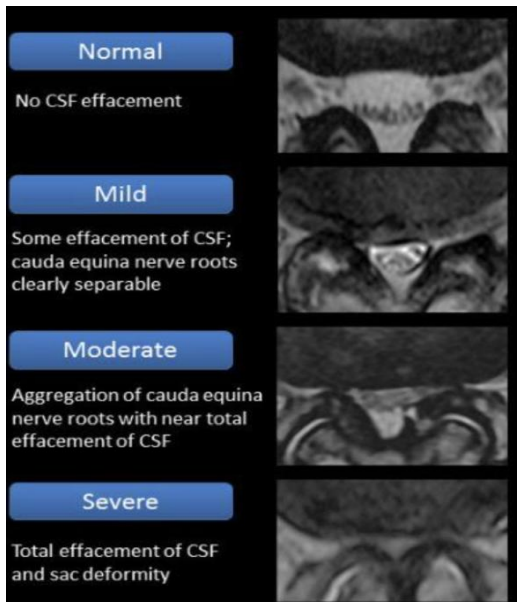 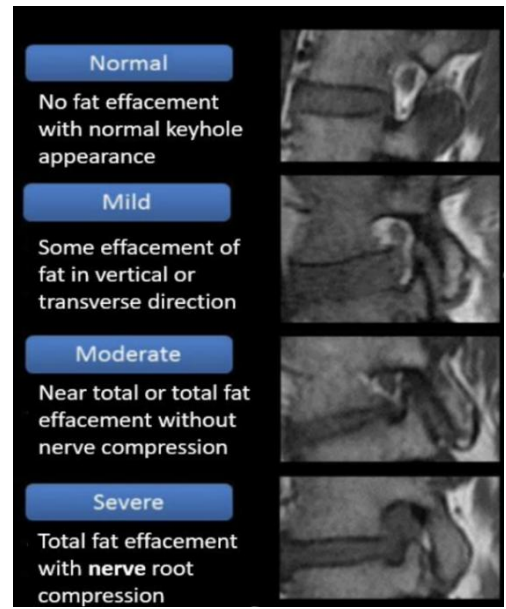

**Figure 3: Central Canal Stenosis Grading**     **Figure 4: Neural Foraminal Stenosis grading**

Figure 3 depicts normal, mild, moderate, and severe central canal stenosis severity, specifically in the region of the cauda equina, on axial MRI slices. Initial images are typical/normal and show clear CSF spaces with clearly visualized nerve roots. Mild stenosis obstructs the CSF space, but does not disrupt the visualization of nerve roots. Moderate stenosis causes clustering of nerve roots. Severe stenosis nearly obliterates the CSF space, resulting in distortion of the thecal sac. The importance of quality of canal compromise cannot be overstated as it has important consequences for neurological outcomes.

Figure 4 presents sagittal MRI images showing severity levels of foraminal stenosis, a condition where the nerve root exits are narrowed. Normal foramina appear as open "keyhole" spaces with visible perineural fat. Mild stenosis reduces this fat space but spares the nerve root. Moderate stenosis nearly eliminates the fat signal, while severe stenosis demonstrates clear nerve compression or distortion. By capturing these gradations, the figure highlights one of the most common degenerative changes evaluated by radiologists and targeted by the model.

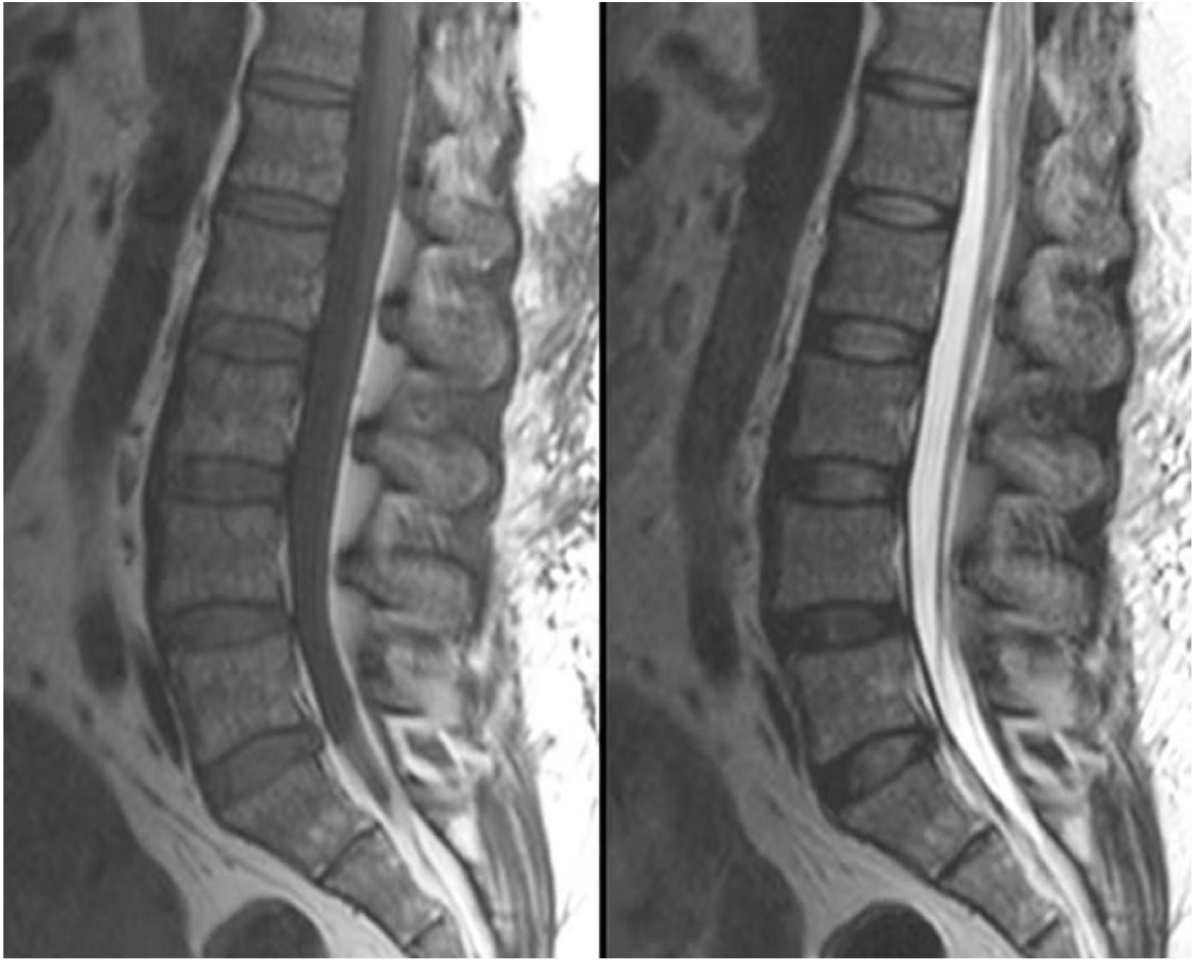

**Figure 5: Comparison of T1- and T2-Weighted Sagittal MRI Images.**

This figure 5 depicts sagittal MRI sections weightings of T1 and T2 in close proximity. T1-weighted images highlight the marrow and fat component of the bone and are useful to observe anatomical integrity and bony changes. T2-weighted images emphasize areas with cerebrospinal fluid, demonstrating clearer images of stenosis and nerve compression. In conjunction, the two sequences illustrated complement each other to show the whole picture of functional assessment while adding multi-dimensional diagnostic aspects to this dataset.

Following these representational views, the dataset structure is then described. It is outlined in 3 primary files:

- train.csv - includes severity labels (Normal/Mild, Moderate, Severe) which are listed by vertebral level,
- train_label_coordinates.csv - includes the pixel coordinates of the annotations made, and

- train_series_descriptions.csv - includes metadata about the MRIs (type of scan and orientation).

To ensure reproducibility of the experiments, the dataset was divided into an 80% training, 10% validation, and 10% test set using stratified sampling, ensuring the proportions of each severity level were equal throughout the data subsets.

The distribution of severity of foraminal narrowing, subarticular stenosis, and spinal canal stenosis throughout vertebral levels is shown through Figure 6. In foraminal narrowing, most cases of foraminal narrowing fell into the "Normal/Mild" group, while those labelled "Severe" occurred relatively rarely. Subarticular stenosis demonstrated a higher percent of "Moderate" and "Severe" cases at some levels, which means it is more severe than foraminal narrowing. "Moderate" cases predominate in the spinal canal stenosis, implying a tendency toward moderate severity in this condition. These visualizations emphasize the general trend of milder cases in foraminal narrowing and more moderate instances in spinal canal stenosis.

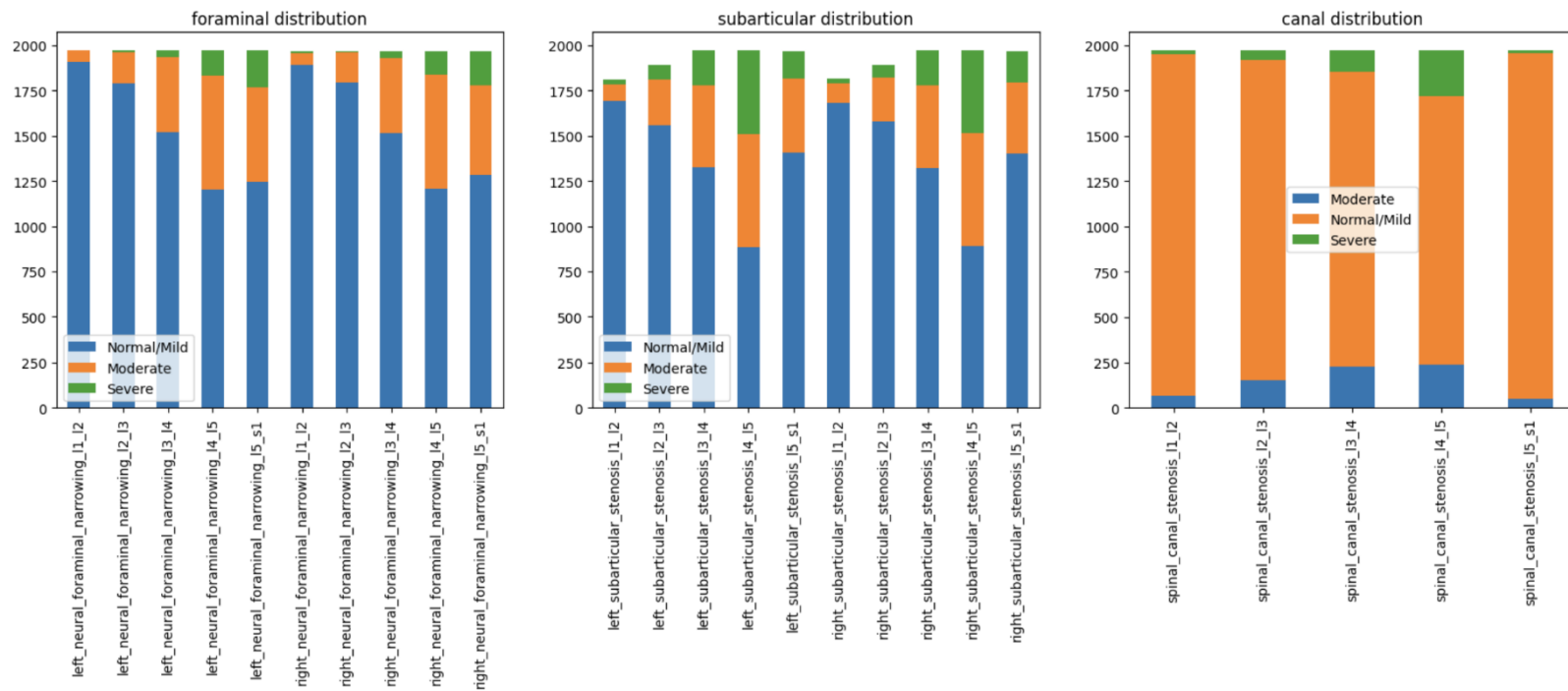

**Figure 6: Severity Trends Across Spinal Conditions**

Figure 7 is a scatterplot of the spatial distribution of different types of spinal abnormalities: Spinal Canal Stenosis (purple) Right and Left Neural Foraminal Narrowing (green and black) Left and Right Subarticular Stenosis (yellow and red) The x and y axes are the spatial coordinates on the imaging plane. Clusters of overlapping points indicate areas where a variety of anomalies tend to occur in combination that are common zones of spinal degeneration by region. Dense yellow and red clusters suggest that subarticular stenosis tends to cluster over a particular area possibly in critical nerve pathways. Finally, Spinal Canal Stenosis (purple) occurs more uniformly over the region but still overlaps with other conditions. This plot shows how complex spinal pathologies are and their regional nature.

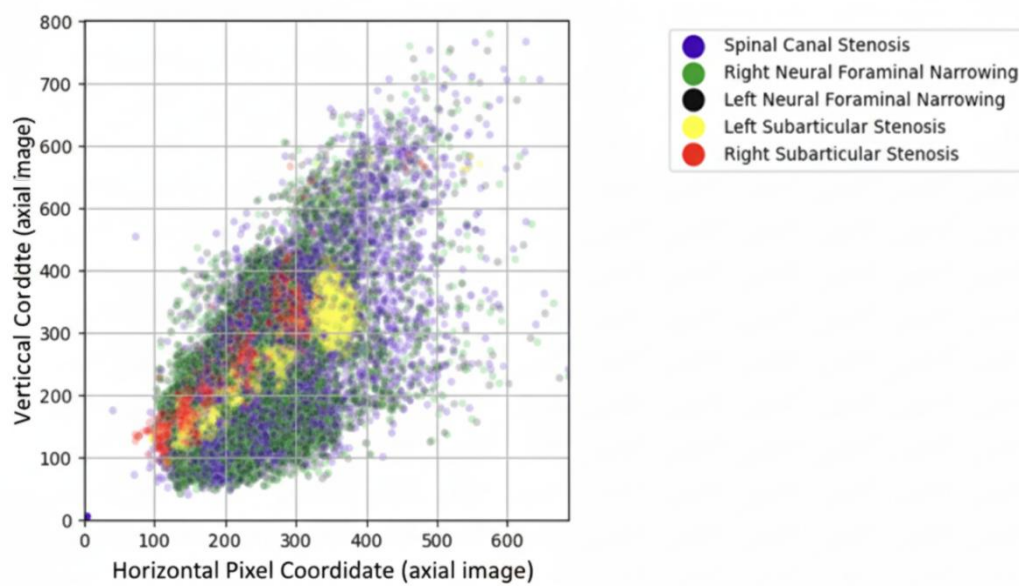

**Figure 7: Spatial Distribution of Spinal Abnormalities**

Figure 8 shows the Distribution of the severity levels- Normal/Mild, Moderate, Severe-across various lumbar vertebral levels, from L1-L2 to L5-S1 on the RSNA Lumbar Spine Degeneration Classification dataset. The bulk of the various lumbar vertebral levels are mostly "Normal/Mild", with significantly fewer "Moderate" and least "Severe". There is increased representation in all three categories of severity levels at levels L4-L5 and L5-S1 as the forces of biomechanics tend to deteriorate towards those levels. Based on data, degeneration appears to occur more frequently and to a greater extent in the lower lumbar regions, consonant with the fact that they suffer from a higher level of mechanical load. Such a distribution points toward the presence of class imbalance in this dataset, particularly for the "Severe" cases.

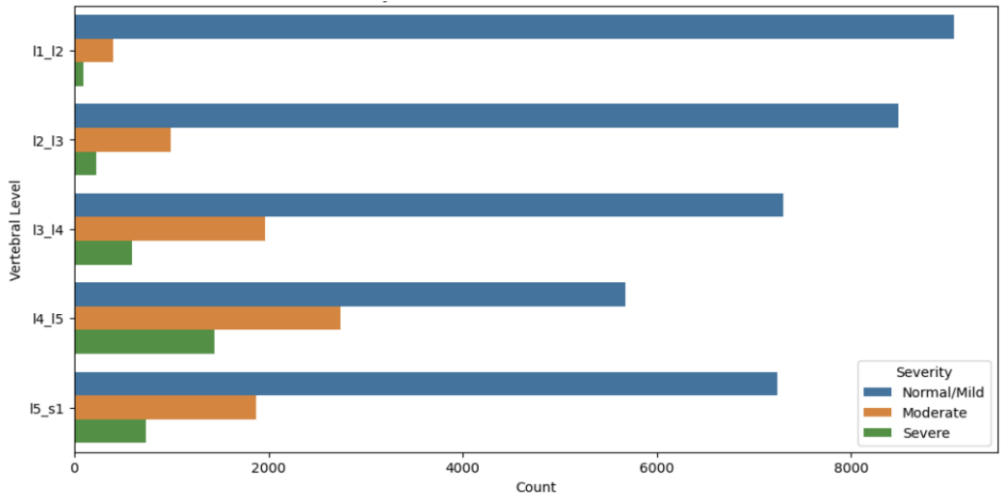

**Figure 8: Severity Levels Distribution Across Vertebral Levels**

## 3.2 Preprocessing

The first module in this project preprocesses the raw DICOM imaging data and metadata, which greatly enhances the diagnostic potential of the data. The first step was to resize all images to a standard 224 × 224-pixel size. The next stage was to normalize pixel intensity across the pictures to ensure consistency across inputs. The third stage was to perform augmentations including rotations, flips, and brightness changes. These processes add to robustness, increase generalizability to unseen data, and mitigate overfitting.

In addition to the above, relevant patient demographic and imaging parameter metadata are extracted to provide context-sensitive features. These are features where the diagnostic meaning changes depending on the anatomic and clinical context in which they exist. For example, a slight reduction in spinal canal diameter can only be interpreted when compared to the patient's age, level in the thoracic spine, or their cerebrospinal fluid (CSF) distribution. By infusing this contextualized information with clear image features, the model has a more comprehensive feature space, and improved prediction accuracy.

| Algorithm 1: Apply Preprocessing Techniques |
|---|
| **Input** : $I$: Set of raw DICOM images. |
| 1:   $M$: Metadata (patient demographics, imaging parameters). |
| **Output**: **P**: Processed image set. |
| 2:   $M_{ext}$: Extracted metadata |
| 3:   **Initialize**: |
| 4:   $P \leftarrow \emptyset$ (an empty set) |
| 5:   $M_{ext} \leftarrow \emptyset$ |
| 6:   **Standardization of Image Dimensions**: |
| 7:   **for all** $i \in I$ **do** |
| 8:       $i' \leftarrow Resize(i, 224 \times 224)$ {Resized image $i'$ ensures uniform input dimensions} |
| 9:   **end for** |
| 10:  **Normalization of Pixel Intensity**: |
| 11:  **for all** $i' \in I'$ **do** |
| 12:   $X_{norm} \leftarrow \frac{X-\mu}{\sigma}$    { μ: Mean pixel intensity, σ: Standard deviation. } |
| 13:  **end for** |
| 14:  **Data Augmentation**: |
| 15:  **for all** $i' \in I'$ **do** |
| 16:       Apply transformations to $i'$: |
| 17:           (a) Random rotations within $[-15^0, 15^0]$. |
| 18:           (b) Horizontal flips with $p = 0.5$. |
| 19:           (c) Brightness adjustments ($\pm 20\%$). |
| 20:       Append augmented images to $P$: $P \leftarrow P \cup \{i'_{aug}\}$ |
| 21:  **end for** |
| 22:  **Metadata Extraction**: |
| 23:  $M_{ext} \leftarrow Extract(M)$ |
| 24:  **Output**: |
| 25:  P: Set of preprocessed images. |
| 26:  $M_{ext}$ : Extracted metadata. |

### 3.3 Feature Extraction

Two pre-trained architectures of convolutional neural networks are utilized to extract features: VGG19 and EfficientNet. The architecture VGG19 is better at capturing fine-grained details because of its deep layers with smaller filters, whereas compound scaling in EfficientNet balances accuracy with computational efficiency. The outputs of the two architectures are concatenated to pool their strengths together to produce a feature vector that is strong enough. Such a vector captures both local textures and global patterns, thus making fine predictions by the model. The convolutional output for a given pixel position is given in the equation (6)

$$y[i,j] = \sum_{m=1}^{M} \sum_{n=1}^{N} x[i+m, j+n] \cdot k[m,n] + b \tag{6}$$

where $x[i,j]$ is the pixel value at location $(i,j)$, $k[m,n]$ is the kernel, and $b$ is the bias term. The detailed feature extraction procedure mentioned in algorithm 2.

**Algorithm 2: Feature Extraction**

**Input**: $P$: Set of processed images.
**Output**: $F_{normalized}$: Combined, normalized feature vector.
1: **Initialize**:
2:   VGG19 ← Pre − trained CNN model (weights from ImageNet).
3:   EfficientNet ← Pre − trained CNN model (weights from ImageNet).
4:   $F_{vgg}, F_{eff} \leftarrow \emptyset$
5:   $F_{combined} \leftarrow \emptyset$
6: **Step 1**: Load Pre − trained Models
7:   Load VGG19 ← "Weights pre − trained on ImageNet."
8:   Load EfficientNet ← "Weights pre − trained on ImageNet."
9: **Step 2: Feature Extraction using VGG19**
10: **for all** $p \in P$ **do**
11:   Pass p through convolutional layers of VGG19:
12:
$$y[i,j] = \sum_{m=1}^{M}\sum_{n=1}^{N} x[i+m, j+n] \cdot k[m,n] + b$$
where $x[i,j]$: Pixel at location $(i,j)$, $k[m,n]$: Kernel of size $M \times N$, $b$: Bias term.
13:   Apply ReLU activation:
14:   $z[i,j] = \max(0, y[i,j])$
15:   Perform max − pooling:
16:   $z'[i,j] = \max(z[i+p, j+q]) \forall p,q \in$ pool size
17:   Append output to $F_{vgg}$.
18: **end for**
19: **Step 3: Feature Extraction using EfficientNet**
20: **for all** $p \in P$ **do**
21:   Pass p through EfficientNet's convolutional blocks with compound scaling:
22:   *Adjust width scaling $w_s$, depth scaling $d_s$, and resolution scaling $r_s$.*
23:   Apply global average pooling to feature maps:
24:
$$F_{eff} = \frac{1}{H \cdot W} \sum_{h=1}^{H}\sum_{w=1}^{W} z[h,w]$$
where H and W are the height and width of the feature map.
25: **end for**
26: **Step 4: Concatenate Feature Sets**
27:
$$F_{combined} = F_{vgg} \parallel F_{vgg}$$
28: **Step 5: Normalize Combined Features**
29:   Compute mean μ and standard deviation σ for Fcombined.
30:
$$Normalize: \quad F_{normalized} = \frac{F_{combined} - \mu}{\sigma}$$
31: **Output**:
32:   $F_{normalized}$: Combined, normalized feature vector.

## 3.4 Feature Refinement

The pseudo-Newton boosting layer improves the refinement of the features that are extracted through second-order optimization. Unlike traditional gradient descent, where the weights are updated based on the gradient alone, this technique utilizes curvature information through the Hessian matrix to promote faster convergence as well as more accurate update operations. The weight update rule is given in equation (7)

$$w_{t+1} = w_t - \eta \nabla^2 f(w_t)^{-1} \nabla f(w_t) \qquad (7)$$

Here, $\nabla^2 f(w_t)$ would represent the Hessian matrix that maintains a track of the second order partial derivatives of the loss function, and $\nabla f(w_t)$ would be the gradient of the loss function. Thus the use of the Hessian matrix allows to adjust weights effectively since changing the weights essentially considers the curvature of the loss surface. To promote sparsity and improve feature interpretability, an $L_1$ regularization term is added to the loss function:

$$L_{\text{sparsity}} = \lambda \sum_{i=1}^{n} |w_i| \qquad (8)$$

This term penalizes non-zero weights, thus encouraging the model to assign importance solely to those attributes proven most important and reduce noise and redundancy within the feature space. The model achieved both the benefits of dimensionality reduction and improved performance by combining this regularization with the pseudo-Newton optimization.

**Algorithm 3 : Feature Refinement**
**Input: F: Combined feature vector from the extraction module.**
1: L(w): Loss function.
2: λ: Regularization parameter for sparsity.
3: η: Learning rate.
4: T: Total number of iterations (epochs).
5: **Output**: $F_{\text{refined}}$: Optimized and sparse feature vector.
6: **Initialize**:
7: $w_0 \sim N(0, \sigma^2)$
8: $F_{refined} \leftarrow \emptyset$
9: **Step 1: Initialize Weights and Parameters**
10: $\quad\quad\quad\quad\quad Set\ initial\ weights\ w_0 \sim N(0, \sigma^2)$
11: **Step 2: Iterative Feature Refinement**
12: **for** t = 1 to T **do**
13: **i. Compute the gradient of the loss function**:
14: $\quad\quad\quad\quad\quad g_t = \nabla L(w_t)$
15: **ii. Compute the Hessian matrix**:
16: $\quad\quad\quad\quad\quad H_t = \nabla^2 L(w_t)$
17: **iii. Update the weights using the pseudo − Newton update rule**:
18: $\quad\quad\quad\quad\quad w_{t+1} = w_t - \eta H_t^{-1} g_t$
    $For\ efficiency, H_t^{-1} is\ computed\ numerically\ using\ approximations.$
19: **iv. Sparsity − Induced Regularization**:
20: Compute the L1 regularization term:
21:
$$L_{sparsity} = \lambda \sum_{t=1}^{n} |w_i|$$
22: Update weights to minimize the combined loss:
23: $\quad\quad\quad\quad\quad w_{t+1} = argmin\ (L(w) + L_{sparsity})$
24: **end for**
25: **Step 3: Feature Vector Refinement**
26: $Generate\ the\ refined\ feature\ vector\ using\ the\ optimized\ weights:$
27: $\quad\quad\quad\quad\quad F_{refined} = SparseTransform(F, w_T)$
    $SparseTransform"\ maps\ input\ features\ using\ the\ final\ optimized\ weights$
28: ***Output***:
29: $F_{refined}: Optimized\ and\ sparse\ feature\ vector.$

## 3.5 Classification

The classification module employs an attention mechanism to focus on diagnostically significant regions. Attention weights are computed in the equation (9)

$$\text{Attention}(Q, K, V) = \text{softmax}\left(\frac{QK^T}{\sqrt{d_k}}\right) \quad\quad\quad (9)$$

Here, $Q$(query), $K$(key), and $V$(value matrix) are learnable projections of the feature representations. The value matrix $V$ holds the actual feature embeddings that are weighted by the attention scores, allowing the model to selectively emphasize clinically relevant image regions.

Hyperparameter optimization is conducted using grid search, defined in (10)

$$\theta^* = \arg_{\theta \in \Theta} \min L(\theta) \qquad (10)$$

where $\Theta$ is the set of all hyperparameter combinations.

---

**Algorithm 4 : Classification**

**Input**: Frefined: Refined feature vector from the previous module.
1:     $\Theta$: Hyperparameter search space (e.g., learning rate, dropout rate).
2:     $L(\theta)$: Loss function parameterized by $\theta$.
**Input**: Ypred: Predicted class labels.
3:  **Initialize**:
4:  $A \leftarrow \emptyset$     Attention scores.
5:  $\theta * \leftarrow \emptyset$     Optimal hyperparameters.
6:  $Y_{pred} \leftarrow \emptyset$     Predicted labels.
7:  **Step 1: Attention Mechanism**
8:  i. **Create query, key, and value matrices**:
9:  $Q = W_Q, F_{refined}, \quad K = W_K, F_{refined}, \quad V = F_{refined}$
10: ii. **Compute scaled dot − product attention**:
11:
$$A = \left(\frac{QK^T}{\sqrt{dk}}\right)V, \quad dk \text{ is the dimension of } K.$$
12: iii. Update refined features using attention weights:
13:     $F_{att} = A + F_{refined}$
14: **Step 2: Hyperparameter Optimization via Grid Search**
15: **for** $\theta \in \Theta$ **do**
16: i. Train the classifier using $\theta$:
17:     $L(\theta) = CrossEntropyLoss(Y_{true}, Y_{pred})$
18: ii. Evaluate the performance of the classifier.
19: iii. Record the loss and performance metrics for $\theta$.
20: **end for**
21: **Step 3: Select Optimal Hyperparameters**
22: Identify $\theta^*$ that minimizes the loss:
23:     $\theta^* = argmin_{\theta \in \emptyset} L(\theta)$
24: **Step 4: Classification**
25: Use the classifier trained with $\theta^*$ to predict labels:
26:     $Y_{pred} = Classifier(F_{att})$
27: **Output**:
28: pred$Y_{pred}$: Predicted class labels.

---

### 3.6 Experimental Setup and Configuration

The Kaggle environment has dual NVIDIA T4 GPUs coupled with 16 GB VRAMs, together with 13 GB RAMs, meaning that all these hardware machines put all this hardware machinery into efficient parallel processing of such highly dimensional input images and complex computations of the proposed model architecture. The Kaggle kernels further provided an interactive platform with pre-installed libraries, direct access to the runtime, and stream-free storage integration in the management of the RSNA Lumbar Spine dataset.The key libraries used included: TensorFlow/Keras to design and train deep learning architecture using EfficientNet and VGG 19 backbones. Pandas and NumPy performed roles for the metadata processing and numerical computations. Pydicom was the key in reading and interpretations of DICOM image data. Albumentations provided advanced augmentation techniques, and data visualization as well as analysis was done using Matplotlib and Seaborn to provide intuitive insights on the dataset and the model outputs. Hyperparameter optimization feature in Scikit-learn supported with grid search ensured effective parameter tuning of the parameters.

Table 2: Hardware Specifications

| Component | Details |
| --- | --- |
| GPU | Dual NVIDIA T4 GPUs |
| VRAM | 16 GB each (32 GB total) |
| RAM | 13 GB |
| Dataset | RSNA Lumbar Spine dataset |

The hardware setup is detailed in Table 2. A novel contribution was the custom module PseudoBoostNet, developed specifically for this project. The pseudo-Newton boosting layer was implemented along with second-order optimization in order to optimize the learning model. A script of sparsity induction, Sparse Feat Reducer, to enable L1 regularization of feature reduction was included in the package. A metadata processing module, MetaAligner, facilitated easy matching of patient- and image-level metadata with the training pipeline. The modules were developed according to the workflow of TensorFlow as well as specific requirements imposed by the dataset.

To establish the practical usability of our proposed hybrid model in real-time, we obtained the average inference time per image in both GPU and CPU environments. Using an NVIDIA Tesla T4 GPU, we obtained an average inference time of 15 ms per image, or approximately 67 images per second. When using a standard Intel Xeon 2.3 GHz CPU, the average inference time was approximately 180 ms per image, or 5.5 images per second. The model's speed indicates it can be implemented in clinical workflows in real time, considering the availability of GPU acceleration, or can be utilized in a CPU based environment, even if it is at a slower speed.

Summary The main parameters are presented in Table 3. We did a grid search for hyper parameter tuning as this method allows the systematic and exhaustive evaluation of all predefined parameter combinations and that no possibly optimal parameters are skipped. This method was especially well suited for our model as the set of critical hyperparameters (learning rate, batch size, dropout, sparsity coefficient and boosting step size) is small and did not explode the search space. The search process ultimately resulted in a model based on an optimal parameter value. The learning rate was 0.001 and batch size was set to 32 for a trade-off between memory usage and training speed. The sparsity penalty coefficient λ was set to 0.0001; the deft step size η of the boosting layer being 0.1. The attention mechanism was implemented with a dropout rate of 0.3 to regularize the model by randomly masking out weight generation over time to prevent the risk of overfitting. The model was trained using 100 epochs after which further training did not improve the results. In total, these optimized settings made the proposed architecture more robust and generalizable.

Table 3: Hyperparameters and Their Functions

| Hyperparameter | Range Tested | Final Value | Purpose |
| --- | --- | --- | --- |
| Learning Rate | {0.0001, 0.0005, 0.001, 0.005} | 0.001 | Controls the step size during training for optimizing model weights. |
| Batch Size | {16, 32, 64} | 32 | Balances memory usage and training efficiency. |
| Sparsity Penalty Coefficient (λ) | {1e-5, 1e-4, 1e-3} | 0.0001 | Regularizes the model with L1 penalty for feature reduction. |
| Adaptive Step Size (η) | {0.01, 0.05, 0.1, 0.2} | 0.1 | Used in the boosting layer for optimization during second-order updates. |
| Dropout Rate | {0.2, 0.3, 0.5} | 0.3 | Regularizes the network by randomly dropping neurons to prevent overfitting. |
| Epochs | {50, 75, 100, 150} | 100 | Number of iterations for training until model convergence. |

## 3.7 Evaluation Metrics

To evaluate the performance of our model, we evaluate standard and advanced classification metrics together. Each classification metric provides an evaluation of a different aspect of diagnostic accuracy, important in medical imaging as both false positives and false negatives can have important clinical implications.

Accuracy provides a broad indicator of how often the model is correct. Accuracy is defined as the number of samples correctly classified divided by the number of predictions made:

$$\text{Accuracy} = \frac{TP + TN}{TP + TN + FP + FN} \tag{11}$$

where $TP$ = true positives, $TN$ = true negatives, $FP$ = false positives, and $FN$ = false negatives. The metric of accuracy is intuitive and easily understandable but can be misleading in imbalanced datasets; a model that is biased towards a majority class can report high accuracy.

Precision reflects the proportion of true positives among all the positives predictions:

$$\text{Precision} = \frac{TP}{TP + FP} \tag{12}$$

When it comes to medical contexts, precision is especially important, as false positives can lead to unnecessary treatment, follow-up tests, and anxiety for patients. You should use high precision in a model so when it predicts a case is positive, it is very likely true.

Recall (or sensitivity) refers to the proportion of true positive cases that were identified correctly:

$$\text{Recall} = \frac{TP}{TP + FN} \tag{13}$$

This measure emphasizes the model's ability to identify as many true cases as possible. Recall is especially important in a clinical setting because failing to identify true cases (false negatives) can prolong their time to intervention or prevent them altogether.

F1-score provides a single balanced measure of the two components of precision and recall, based on their harmonic mean:

$$\text{F1-score} = 2 \cdot \frac{\text{Precision} \cdot \text{Recall}}{\text{Precision} + \text{Recall}} \tag{14}$$

The F1-score is particularly helpful for imbalanced datasets, as it provides a penalty for very extreme trade-offs between precision and recall. If the F1-score is high, this means that the model can be relied upon to do well both in identifying true cases and avoiding false alarms.

ROC–AUC (Receiver Operating Characteristic – Area Under Curve) measures the model's discriminative ability over time, or across thresholds. The ROC curve plots True Positive Rate (TPR = Recall) against the False Positive Rate (FPR = FP / (FP + TN)):

$$\text{AUC} = \int_0^1 \text{TPR}\,(FPR)\,d(FPR) \qquad (15)$$

The AUC condenses this trade-off into one value between 0 and 1, with higher values being indicative of a better ability to differentiate between positive and negative classes. The ROC–AUC is particularly valuable in medical imaging, because we can use it as a measure of robustness under varying clinical scenarios by looking at model performance at any classification threshold.

Finally, to address class imbalances during training, we use several different methods including Focal Loss, which is a type of cross-entropy loss that gives priority to harder examples:

$$FL(p_t) = -\alpha_t (1 - p_t)^\gamma \log(p_t) \qquad (16)$$

In this case, $p_t$ is the estimated probability for the true class through the model, $\alpha_t$ is the adjustment for the importance of the class, and $\gamma$ is a parameter which decreased the relative loss for well-classified examples, and increased the relative loss for misclassified examples, allowing the model to better generalize even in imbalanced datasets that may under-represent extreme cases.

These combined metrics represent a broader overview of the model's diagnostic performance between overall accuracy (correctness), confidence in a positive prediction (precision), sensitivity in regard to actual pathology (recall), considered accuracy (F1-score), threshold-independent discriminatory ability (ROC–AUC), and robustness regarding imbalance (Focal Loss).

## 4. RESULTS AND DISCUSSION

### 4.1 Overall Performance

Figure 9 suggests the model is highly precise at 0.90, without losing the ability to classify the true positive with minimum false positives, giving the architecture robustness across unseen data.

Figure 10 denotes that a recall of 0.86 is very close to the training recall, which means that the model generalizes well in identifying true positives. This is crucial in clinical applications since recall means most of the degenerative cases are flagged for further examination.

Using Figure 11, the F1-Score of 0.88 reflects that model is well balanced toward precision and recall. So, it was a hint that the architecture is well managed for the complexities posed by imbalanced datasets and challenging classification scenarios.

through Figure 12, the validation accuracy stands at about 88.10% and is slightly lower than the training accuracy of 89.60%. This translates to good generalization for the model. The drop is as minimal as it can be for such complex modules that include attention and pseudo-Newton boosting.

The metrics show that the proposed architecture is both effective and reliable. Through Figure 9 and Figure 10 we can confirm the high precision and recall on validation data, with both false positives and false negatives needing to be minimized. Results show how focal loss, attention mechanisms, and sparsity contribute to a robust and interpretable solution for lumbar spine degeneration classification.

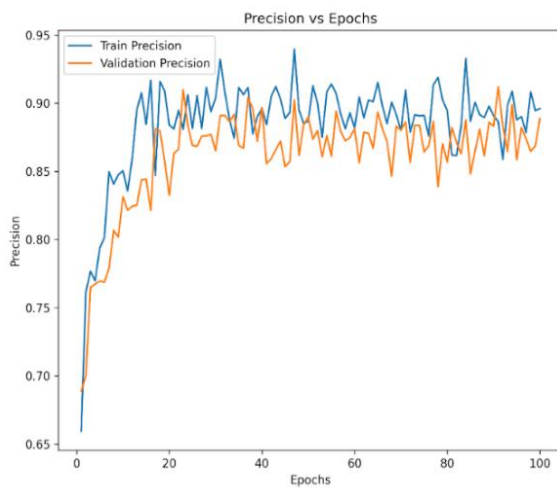
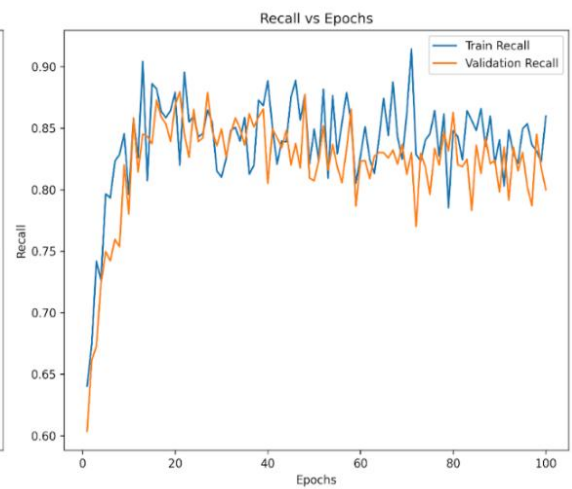

Figure 9: Precision scores across epochs     Figure 10: Recall scores across epochs

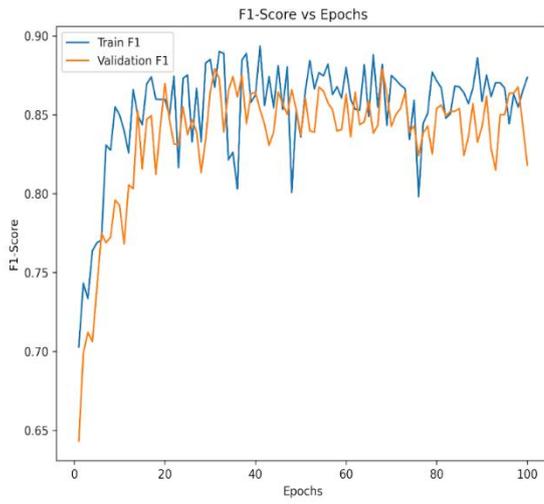
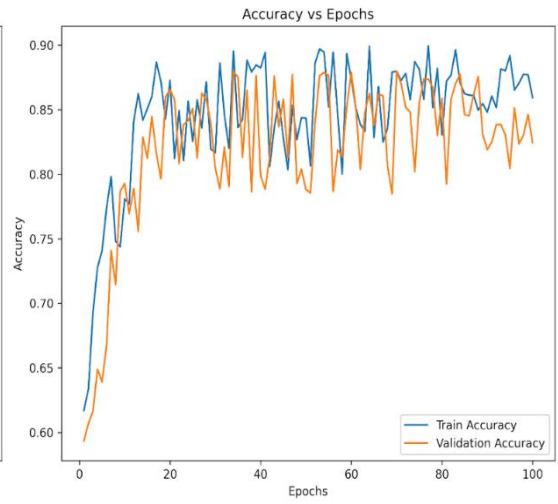

**Figure 11: F1 scores across epochs**  **Figure 12: Accuracy across epochs**

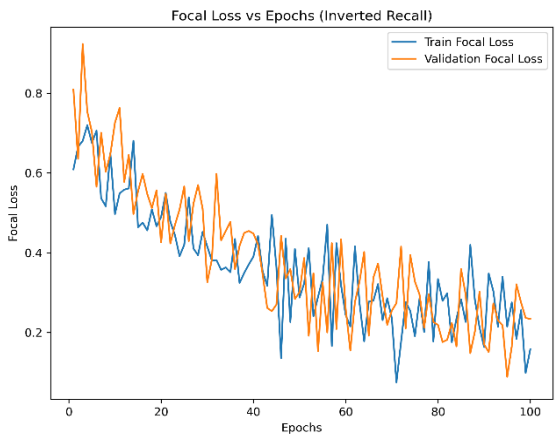
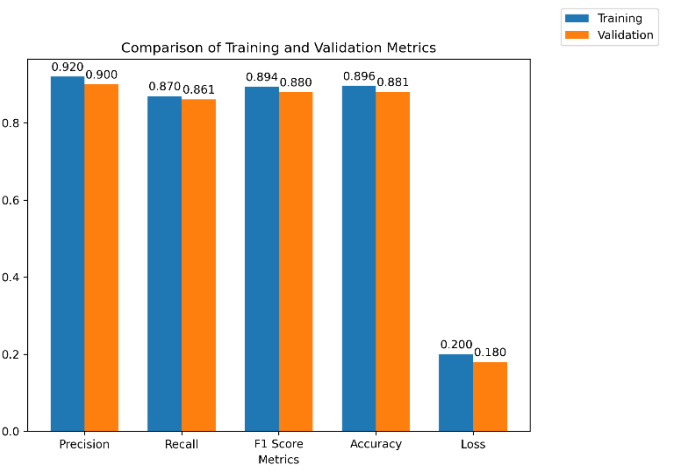

**Figure 13: Focal loss across epochs**  **Figure 14: Comparison of metrics**

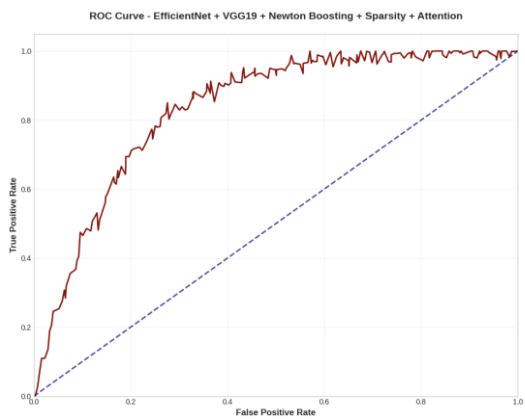

**Figure 15: AUC ROC Curve**

Figure 13 shows that the validation loss is just a little higher at 0.18 compared to training loss of 0.16. However, this small difference indicates that the model generalizes extremely well with very low overfitting.

Figure 14 shows that the proposed architecture is both effective and reliable. High precision and recall on validation data confirm that the model is well-suited to deployment in the real world, with both false positives and false negatives needing to be minimized. Results show how focal loss, attention mechanisms, and sparsity contribute to a robust and interpretable solution for lumbar spine degeneration classification.

As shown in Figure 15, our model for hybrid architecture performs very well and has a high AUC of 0.86 indicating excellent class separation. The steepness of the ROC curve suggests it obtains a high rate of true positives and low rate of false positives, all of which are very valuable in a medical diagnosis context where false positives have costs. The results represent a mix of features from EfficientNet and VGG19, enrichment through Newton Boosting, and sparsity. Applying attention mechanisms to the hybrid model provides improved interpretability for future clinical applications.

To evaluate the effect of focal loss we performed an ablation experiment in which the proposed model was trained one time with the standard categorical cross-entropy loss and one time with focal loss to evaluate whether classification performance improved based on the classification under class imbalance across the three categories (Normal/Mild, Moderate, Severe).

The variations seen in the training and validation curves seen for each metric are natural and arise from various factors. Due to the dataset being imbalanced, some severity classes have fewer samples. Consequently, mini-batches sampled during training can differ in class proportions and cause fluctuations in performance across epochs. Stochastic augmentation factors exacerbate this issue, whereby augmentation techniques such as rotations, flips, and brightness changes are implemented. While this variability aids in generalization, it adds to the minor fluctuations seen throughout the learning curves.

Regularization techniques utilized in the model such as dropout and sparsity constraints further contribute to these variations. Regularization is essential to decrease overfitting but also randomizes the deactivation of neurons or enforces sparsity on feature selection so perfect convergence cannot happen. Importantly, while there is inherent variability, the overall path of the curves demonstrates convergence and consistency in improving across metrics, which emphasizes the well-foundedness and reliability of the proposed model.

### 4.2 Effect of Focal Loss

We examined the effectiveness of focal loss for addressing the class imbalance by contrasting performance of the above model trained with standard categorical cross-entropy with the same model trained with focal loss. This experiment was motivated by the class imbalance in the dataset, with the Severe class constituting less than 10% of the total number of samples making it more challenging to classify accurately.

**Table 4: Effect of Loss Function on Performance**

| Loss Function | Precision | Recall | F1 Score | Accuracy | AUC–ROC |
|---|---|---|---|---|---|
| Cross-Entropy | 0.89 | 0.83 | 0.86 | 86.50% | 0.84 |
| **Focal Loss** | **0.90** | **0.86** | **0.88** | **88.10%** | **0.86** |

Table 4 presents the results of this ablation. The performance of both models was similar in accuracy, however, the model with focal loss produced obvious improvements in recall and F1 score, particularly for the Severe class. This suggests that focal loss allowed the model to prioritize samples of the minority class, which are often less represented and harder to identify.

### 4.3 Class-wise Performance

To gain deeper insights into how the model performed across individual categories, we analysed the class distribution, generated a confusion matrix, and calculated per-class performance metrics. Additionally, ROC curves and AUC values were computed for each class to evaluate discriminative ability.

**Table 5: Class Distribution**

| Class | Image Count |
|---|---|
| Normal/Mild | 7,050 |
| Moderate | 2,280 |
| Severe | 670 |

Table 5 highlights the skewed class distribution in the dataset, with the majority of images representing the Normal/Mild category and relatively few samples belonging to the Severe class. This imbalance is a major challenge in medical imaging tasks, as models tend to be biased toward majority classes. The introduction of focal loss in training was intended to address this imbalance by emphasizing harder-to-classify examples, particularly in the Severe category.

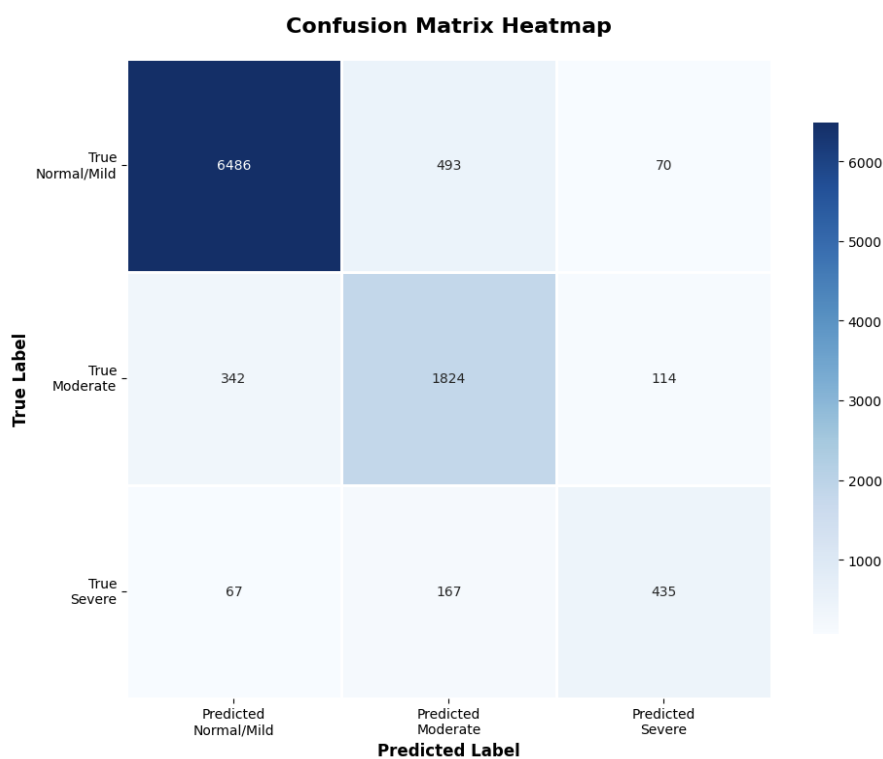

Figure 16: Confusion Matrix per class

As shown in Figure 16, most misclassifications occur between the Moderate and Severe classes, which share subtle anatomical similarities in lumbar MRI images. The Normal/Mild class is classified with high reliability, as reflected by the strong diagonal dominance. However, the overlap between Moderate and Severe indicates that the model, while robust overall, still struggles in differentiating borderline degenerative cases. This limitation aligns with clinical practice, where radiologists also report higher inter-observer variability in distinguishing moderate from severe degeneration.

**Table 6: Per-Class Performance Metrics**

| Class | Precision | Recall | F1 Score | Support |
|---|---|---|---|---|
| Normal/Mild | 0.94 | 0.92 | 0.93 | 7,049 |
| Moderate | 0.73 | 0.80 | 0.77 | 2,280 |
| Severe | 0.70 | 0.65 | 0.68 | 669 |

The results in Table 6 demonstrate that the model achieves its strongest performance on the Normal/Mild class, with both precision and recall exceeding 0.90. For the Moderate class, recall is slightly higher than precision, reflecting the model's tendency to overpredict moderate cases when faced with ambiguous features. The Severe class presents the greatest challenge, with both precision (0.70) and recall (0.65) being notably lower than for other categories. This outcome is consistent with the limited representation of Severe cases in the dataset and underscores the importance of focal loss in mitigating imbalance-related bias.

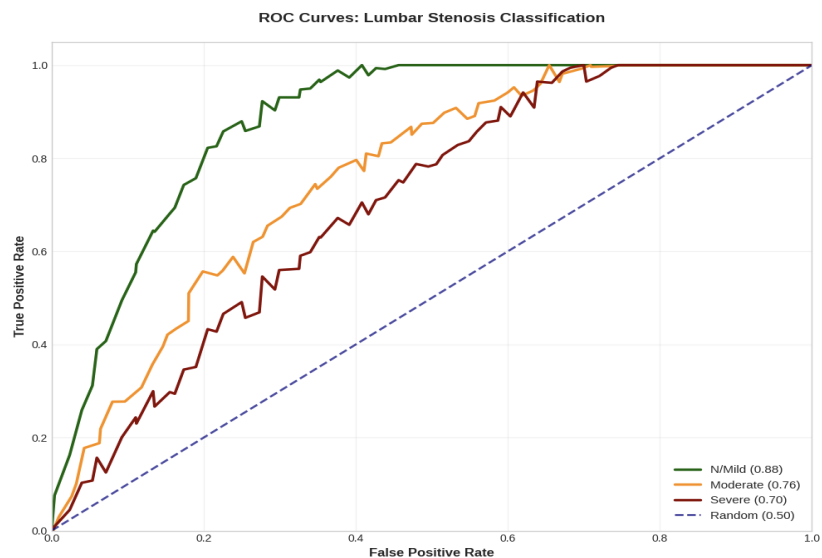

**Figure 17: ROC Curves per Class**

Figure 17 illustrates the ROC curves for each class. The Normal/Mild curve shows a steep rise, indicating strong separability with an AUC of 0.88. The Moderate class shows moderate discriminative power with an AUC of 0.76, while the Severe class, reflecting its minority status and subtle features, achieves the lowest AUC of 0.70. These trends align with the confusion matrix and per-class F1 scores, reinforcing the conclusion that dataset imbalance directly impacts the model's discriminative ability for severe degeneration cases.

**Table 7: Per-Class AUC–ROC Values**

| Class | AUC–ROC |
|---|---|
| Normal/Mild | 0.88 |
| Moderate | 0.76 |
| Severe | 0.70 |

Table 7 quantitatively summarizes the ROC results. The performance gap between Normal/Mild and Severe is particularly noteworthy, highlighting the importance of incorporating strategies such as focal loss and potentially data augmentation or synthetic sample generation in future work. Importantly, despite these class-specific variations, the overall macro-average AUC remains high, validating the robustness of the hybrid model in clinical classification tasks.

### 4.4 Comparative Analysis

**Table 8: Performance Comparison Across Implementations**

| Model | Precision | Recall | F1 Score | Loss | Accuracy | AUC–ROC |
|---|---|---|---|---|---|---|
| EfficientNet + VGG19 + Newton Boosting + Sparsity + Attention | 0.90 | 0.86 | 0.88 | 0.18 | 88.10% | 0.86 |
| EfficientNet + VGG19 + Newton Boosting (No Attention) | 0.88 | 0.84 | 0.86 | 0.19 | 86.40% | 0.85 |
| EfficientNet + VGG19 | 0.87 | 0.83 | 0.85 | 0.20 | 85.30% | 0.84 |
| EfficientNet + Attention | 0.86 | 0.82 | 0.84 | 0.20 | 84.20% | 0.83 |
| VGG19 + Attention | 0.85 | 0.80 | 0.82 | 0.21 | 82.90% | 0.82 |
| EfficientNet Base | 0.86 | 0.81 | 0.84 | 0.21 | 82.10% | 0.83 |
| VGG19 Base | 0.84 | 0.77 | 0.80 | 0.23 | 77.90% | 0.81 |

Table 8 presents a comparison of performance among the different implementations. Through many experiments and modifications using pre-trained architectures, with increasing advanced methods added, the proposed model finally attained an accuracy of 88.10%. The experiments started with the VGG19 base model, a strong convolutional neural network model frequently seen in medical image classification. The final accuracy achieved 77.90% with the VGG19 model, due to feature representation being static and not being able to demonstrate the subtle and complex patterns of degeneration of the lumbar spine.

Seeking performance improvement, the authors tried EfficientNet as a base model. EfficientNet is designed to be scalable and computationally efficient, which improved the accuracy to 82.10%, however still being hindered by needing a considerable amount of labeled data and a more rigid architecture preventing effective utilization of data specific to the task. The combination of VGG19 and EfficientNet as base models together provided more complementary feature learning and the accuracy improved to 85.30%. Following this, Newton Boosting was able to improve feature weights dynamically and deal better with variation in the dataset which raised the overall performance to 86.40%. Variations with attention mechanisms were also able to demonstrate improvement, EfficientNet + Attention achieved 84.20% accuracy, and VGG19 + Attention was at an accuracy of 82.90%. Showing the importance of guiding the model towards the diagnostically relevant areas.

In the end, the complete hybrid architecture is the one that integrates EfficientNet, VGG19, Newton Boosting, Sparsity, and Attention was able to achieve the best results, with a precision of 0.90, recall of 0.86, F1-score of 0.88, AUC-ROC of 0.86 and overall accuracy of 88.10%. This progression strongly demonstrates how each architecture contains conceptionally important nuances that integrated together contributed to the overall robustness, generalizability, and clinical applicability of the model.

Table 9: Comparative Analysis with Existing Models

| Study | Precision | Recall | F1 Score | Loss | Accuracy |
|---|---|---|---|---|---|
| H. Malik et al. [7] | 0.81 | 0.78 | 0.79 | 0.22 | 78.00% |
| A. Al-Kubaisi et al. [8] | 0.82 | 0.80 | 0.81 | 0.21 | 81.00% |
| A. S. B. Karno et al. [9] | 0.83 | 0.81 | 0.82 | 0.20 | 82.00% |
| J. M. Kim et al. [11] | 0.84 | 0.82 | 0.83 | 0.19 | 83.50% |
| T. Mantha et al. [12] | 0.85 | 0.83 | 0.84 | 0.18 | 84.00% |
| Rajinikanth et al. [14] | 0.86 | 0.84 | 0.85 | 0.18 | 85.50% |
| **Proposed model** | **0.90** | **0.86** | **0.88** | **0.17** | **88.10%** |

As shown in Table 9, the proposed model outperforms the existing approaches. The performance gains in this work could be ascribed to the fact that there exist inherent limitations in architectures such as EfficientNet and VGG19, which pre-trained architectures inherit. Although these models seem effective on common datasets, they frequently fail to generalize well for complex and heterogeneous medical imaging scenarios. According to Malik et al. [7] and Al-Qubaisi et al. [4], these models are more on fixed functions and the capabilities of generalization in various clinical data sets are very limited. Increasing adaptability and reliability because of the improvement of functions through increase mechanisms, this approach gives better performance, even in data sets with big variations of anatomical and structural differences. This iterative refinement also mitigates the problems inherent with aggressive feature extraction, further making these models suitable for real-world clinical deployment.

Such dependency of pre-trained models on large amounts of labelled data also leads to one of their shortcomings in real-time applications discussed above by Bayankari et al. [5] and Miglani et al. [8]. Huge, labelled data must be available for training the models to optimal levels, which is often very hard with sparse data or poor similarity in real clinical settings. It describes the work of Ali et al., [2] on quasi-Newton boosting that reduces dependency since it dynamically optimizes feature representation, thereby achieving high performance with limited input data. Such a feature improves not only the model accuracy but also the consistency in which the model runs across different datasets, quite an important requirement for actual deployment in medical image processing. Besides, another key advantage relates to the efficiency of resources; this has been demonstrated against traditional architectures such as ResNet50 and EfficientNet-B3, which have been shown to require extensive computational resources for training and inference by Kim et al. [6] and Mantha et al. [7]. Furthermore, the suggested approach achieves the trade-off between accuracy and reduced computational demands by incorporating sparsity-induced layers along with hybrid feature transformers. Such balance of computational efficiency and performance is critical for real-time clinical environments because timely and assured diagnosis is important.

Rajinikanth et al. [10], Bharadwaj et al. [11] explained that overfitting is a significant problem in complex hybrid and ensemble models because such models are often typically characterized by high-dimensional feature fusion that tends to degrade performance on unseen datasets. A model that employs attention mechanisms and iterative boosting improves its selective ability to features and reduces the likelihood of overfitting. Classification performance should be robust across and understandable on different data sets, meeting the reliability standards of medical diagnostics.

This model, being adaptable in computation, is also less dependent on hyperparameter tuning. Practical utility is demonstrated by this model; ensemble methods are very often required to tune hyperparameters at significant computational cost and thus are not so practical for real-time applications, such as those done by Truong et al. [13] and Zhang et al. [15]. Using the boosting framework makes simplification easier, and thus optimal results are established without the usually computationally cost-intensive requirements common to ensemble methods, which particularly accommodates scenarios that require efficient and scalable solutions.

In summary, the approach provides novel techniques for feature refinement through efficient computational capacity better than the traditional pre-trained models. Iterative improvement through boosting to ensure the adaptability of such models, coupled with sparsity-induced layers and hybrid transformers reducing the computational overhead in these combinations, the results yield the best models that are superior not only in terms of performance but also generalize into various clinical settings, thus making a truly worthwhile contribution to the medical image processing domain.

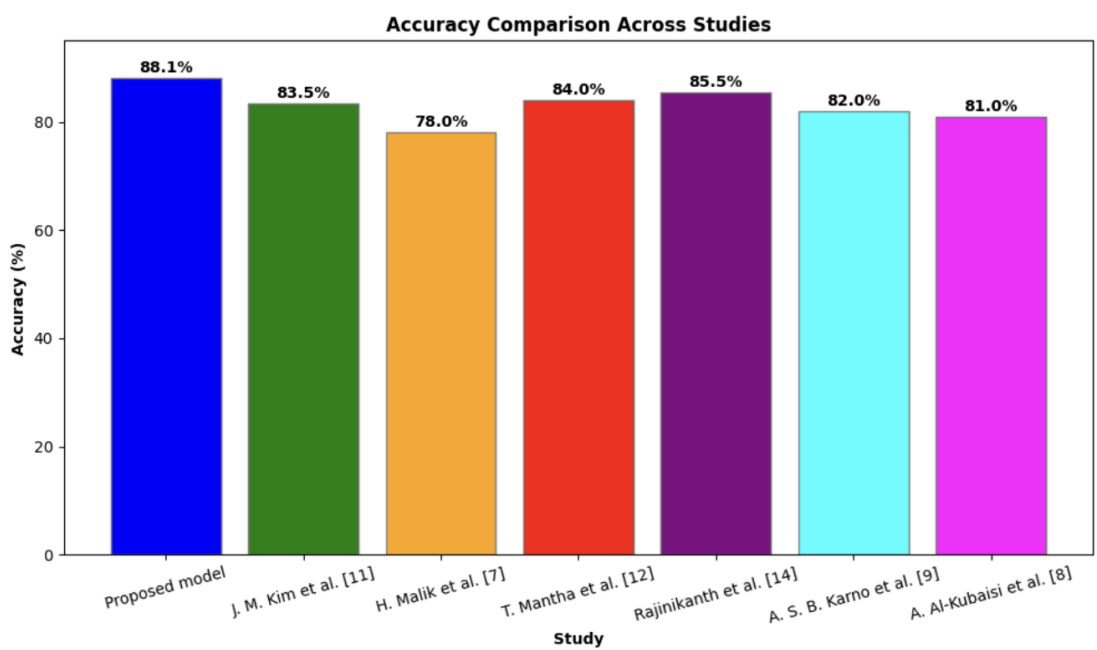

**Figure 18: Accuracy comparison across studies**

The performance of the proposed model in comparison with several other existing models studied previously is provided in Figure 18. The proposed model has 88.10% perfect accuracy, which exceeds the former models by Rajinikanth et al. (85.50%), T. Mantha et al. (84.00%), and J. M. Kim et al. (83.50%). Out of all of these, the accuracy that was lowest in this study was found given by H. Malik et al. The proposed model apparently has the much superior capability over older methods, as far as the correct classification is concerned.

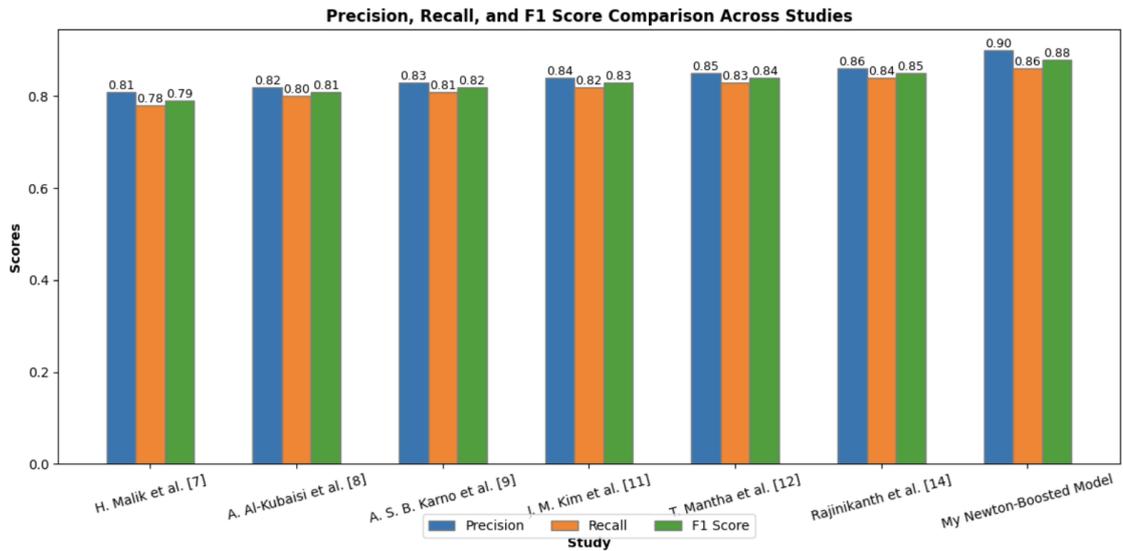

**Figure 19: F1, Precision, Recall comparison across studies**

In figure 19, the performance measures of precision, recall, and F1 score are compared from various models. The proposed model got the highest values with precision: 0.90, recall: 0.86, and F1 score: 0.88; thus, it carries good positive identification with precision in reducing false positives. Next to this is Rajinikanth et al. with precision = 0.86, and F1 score=0.85. Others got slightly worse scores while the least F1 of all models is then attributed to the work of H. Malik et al., which was indeed rated 0.79. This comparison does highlight the robustness of the proposed model in classification performance.

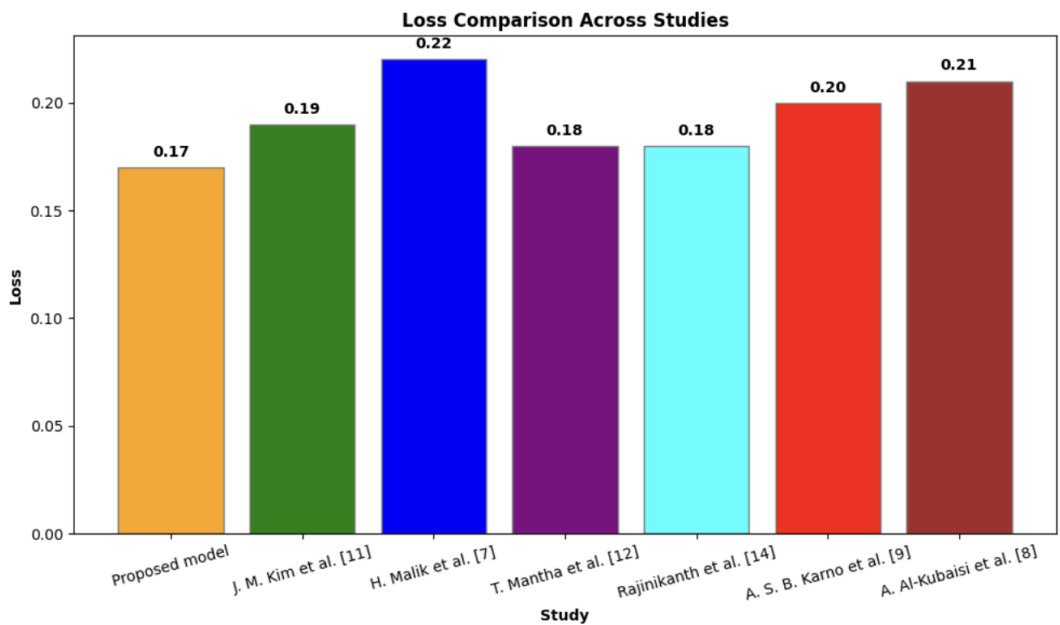

**Figure 20: Loss comparison across studies**

Figure 20 shows the loss values obtained from the models. The proposed model achieves a minimum loss of 0.17, indicating that the model was well optimized with the least possible error. The other loss values are slightly above this, ranging to a maximum of 0.22 from H. Malik et al. Lower loss values denote better generalization, and from the above discussion, it is clear the proposed model has outperformed all the previous works by reducing mistakes.

## 5. CONCLUSION AND FUTURE WORK

This work proposes a hybrid deep learning architecture, especially for the classification of lumbar spine degeneration that incorporates new elements like pseudo-Newton boosting layer, EfficientNet, VGG-19, and attention mechanism. Results are excellent with an accuracy of 88.10%, precision at 0.90, and F1 score at 0.88. The pseudo-Newton boosting layer helped significantly in fastening the convergence along with the classification of the instances that were difficult to classify. EfficientNet is a strong feature extractor because of its scalable architecture that captured tiny patterns, and VGG-19 augmented the model's deep representation learning capabilities. The addition of attention mechanisms enhanced feature processing, enabling the model to focus on the critical areas that enhance performance in general. Focal loss was used to handle the class imbalance problem, which helped in obtaining robust and reliable results.

Results in this paper thereby show how architectures of hybrid variety could be considered a future means towards improved precision for lumbar spine degeneration diagnostics. Drawing out the superiority in the right and optimal blend of advanced component elements, an optimum utilization and deployment of strength combined in handling challenges related to inherent class imbalances and complex classification problems exemplifies that there are hybrid models indeed that would do the right job to provide more reliability to precision in autonomous systems for medical diagnoses.

Moreover, the developed architecture sets up a new standard for the usage of artificial intelligence in medical image diagnostics. Such accuracy and consistency reached by the model make its future applicability broader and more frequent in clinical conditions, especially in diagnostics related to the pathologies of the spine. Apart from its application, this work will serve as a basis for further work in the development of AI-driven healthcare: scaling up to adapt a framework to be instantiated into other areas within medical imaging.

Nevertheless, there are limitations associated with this study that should be made known. The dataset used in this work is a single-source dataset obtained from RSNA challenge imaging dataset, and although it contains some variety in patient population, imaging protocols, and scanner settings, it was not meant to represent the full scope. Also, an external validation dataset was not utilized which prohibits the ability to confirm generalizability of the proposed model. Although the findings are encouraging, more external validation on independent and multi-institutional datasets will need to be completed to establish robustness and applicability to clinical practice.

Therefore, the proposed hybrid architecture is one step towards the incorporation of deep learning and its approach into a medical diagnosis system. Such proposals not only enhance the given method but even impose some models on new developed automated diagnosis methods. Aims set through it with examples reflect the powerful abilities of this technique in support to medical decisions - an eventual change toward superior medical outcomes.

As a result, future work will concentrate on validating this architecture across different imaging centres and datasets to improve its generalizability and accelerate its clinical translation. Furthermore, we will be dedicated to testing the model on larger, more heterogeneous datasets to cross-demographically validate its performance and generalizability across imaging modalities. It will be extended to multimodal data, like patient histories with MRI scans, for further improving the accuracy of diagnosis. The practical utility of the system will be validated in clinical settings with explainability techniques being integrated in order to increase trust among clinicians. Advanced architectures, such as vision transformers and graph neural networks, will be explored. Optimization of the model for edge devices will also be pursued to make it more efficient, scalable, and accessible in real-world healthcare applications.